\documentclass[]{TEAI}

\usepackage{xurl}
\usepackage{booktabs}
\usepackage{graphicx}
\usepackage[export]{adjustbox}
\usepackage{array}
\usepackage{multirow}
\usepackage{makecell}
\usepackage{listings}
\usepackage{float}
\usepackage{nicefrac}

\title{LIBERO-Agent: Evaluating General-Purpose Agents for Direct Embodied Manipulation}

\author{
    Zijie Diao\textsuperscript{1},
    Yitong Chen\textsuperscript{1,2},
    Sicheng Xie\textsuperscript{1,2},
    Tianyi Lu\textsuperscript{1,2},
    Wujian Peng\textsuperscript{1,2},\\
    Guojin Zhong\textsuperscript{1},
    Houze Xu\textsuperscript{1},
    Ziyi Ye\textsuperscript{1},
    Zuxuan Wu\textsuperscript{1,2},
    Yu-Gang Jiang\textsuperscript{1}
}

\affiliation{
$^1$\mbox{Institute of Trustworthy Embodied AI, Fudan University}\\
$^2$\mbox{Shanghai Innovation Institute}
}

\abstract{General-purpose agents can plan, use tools, and revise their behavior from feedback, but it remains unclear whether these capabilities transfer from digital environments to embodied manipulation. To investigate this question, we introduce \textbf{LIBERO-Agent}, an agent-native benchmark for evaluating these agents in robot manipulation tasks. Rather than asking agents to submit task-level Python control programs or operate through high-level robot skills, LIBERO-Agent provides an interactive robotic environment where agents can select which observations to inspect, process them with their own tools, and issue native action commands. LIBERO-Agent integrates 200 tasks into a common interaction framework and provides a 30-task primary suite that separates perception, short-horizon execution, and long-horizon composition. Results reveal a pronounced reliability gap: while agents perform well on perception and easy short-horizon tasks, their performance degrades substantially on hard short-horizon and long-horizon tasks. Richer observations improve short-horizon manipulation, while demonstration benefits depend on the agent and format. Among these agents, GPT-6 Astra achieves the strongest overall performance. Further analysis shows its major advantage lies in mechanism interaction, especially when sustained physical contact is needed, while its remaining failures stem from cross-stage interference and geometric errors.}

\checkdata[Code]{\url{https://github.com/dzj441/Libero-Agent}}

\begin{document}
\maketitle

\section{Introduction}
Large language models have evolved from conversational systems~\citep{ouyang2022training} into general-purpose agents that use tools and carry out complex tasks. Across mathematical reasoning~\citep{cobbe2021training}, code generation~\citep{jimenez2024swe,huang2026deepswe}, web browsing~\citep{zhou2024webarena,koh2024visualwebarena}, and computer use~\citep{xie2024osworld,rawles2025androidworld}, agents can plan toward a goal, revise their plans from feedback, and keep using tools over long interactions. These advances call for more demanding evaluation settings that test whether agents can autonomously organize perception, reasoning, and action throughout an interactive task.

\begin{figure}[t!]
    \centering
    \includegraphics[width=\linewidth]{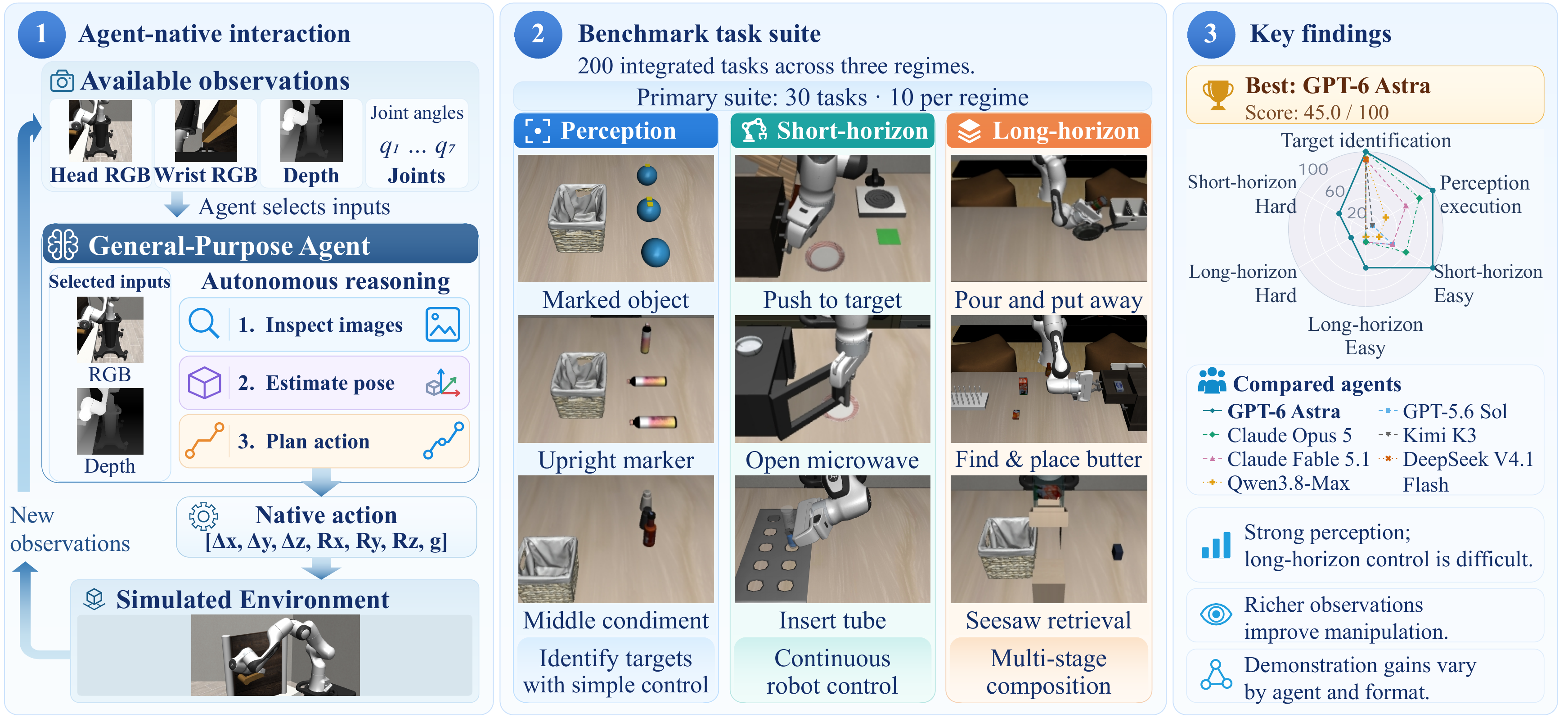}
    \caption{Overview of LIBERO-Agent: agent-native interaction, benchmark task suite, and key findings.}
    \label{fig:overview}
\end{figure}

Embodied robot manipulation~\citep{liu2023libero,james2020rlbench,chen2025robotwin,chen2026robodojo} is a natural setting for such evaluation. An agent must ground instruction-relevant objects in the scene and decide further actions from visual and physical feedback. It must also handle potential occlusion and execution errors while tracking a changing task state under a time budget. Simulation makes these interactions reproducible and scalable. 
However, existing manipulation benchmarks often do not expose all of these choices to the evaluated agent: observations may be preselected, or physical execution may be abstracted behind benchmark-defined task-level action commands. When adapted to general-purpose agents, such scaffolding can decide what information the agent receives or how actions are composed, making it difficult to determine which embodied capabilities come from the agent itself rather than from the benchmark interface.
Moreover, existing benchmarks are rarely structured to diagnose agent capabilities across perception, short-horizon execution, and long-horizon composition, making it difficult to localize where failures arise.

We define \emph{agent-native embodied evaluation} as a setting that makes these choices part of the agent's responsibility: the benchmark specifies the available observations, native robot-control interface, and resource budgets, while leaving observation selection and processing, task decomposition, and action composition to the agent. To support this setting, we adapt LIBERO~\citep{liu2023libero}, a widely used simulation benchmark for robot manipulation, through a minimal interface that exposes episode control and bounded sequences of native end-effector commands. Agents independently choose which available observations to inspect, process them using their own tools, and compose actions without predefined task-level primitives. All interactions are subject to fixed action and time budgets. 

Leveraging LIBERO's compatibility with related manipulation environments, we integrate representative tasks from several benchmarks into a common framework comprising \textit{200} tasks. From this collection, we select 30 representative tasks as our primary evaluation benchmark and separate them into three aspects of embodied competence: perception, short-horizon manipulation, and long-horizon manipulation.
The perception tasks focus primarily on target identification and therefore use only deliberately simple actions. They cover diverse perceptual signals involving object identity, appearance, state, spatial relations, and physical properties.
The short-horizon tasks minimize perceptual ambiguity and long-term planning demands to focus on manipulation execution. Example tasks include pick-and-place, pushing, button pressing, knob rotation, and articulated-fixture manipulation.
The long-horizon tasks compose these abilities across multiple stages, adding requirements for instruction decomposition, scene exploration, cross-stage constraints, and recovery from intermediate failures.

Using this benchmark, we compare seven general-purpose agents under a common protocol and study how different observation inputs and demonstration context affect manipulation. GPT-6 Astra achieves the highest Performance Score of 45.0/100. It succeeds on all perception and easy short-horizon tasks, but drops to
40\% on hard short-horizon manipulation and 22\% stable stage completion
on hard long-horizon tasks. Richer observations improve short-horizon manipulation across Astra, Opus, and Fable, while demonstration benefits depend on both the agent and the representation format. Further analysis shows that Astra's advantage is strongest in mechanism interaction and is associated with sustained physical contact.
Once local execution improves, failure modes shift toward cross-stage interference and geometric errors.

Our contributions are threefold:

\begin{itemize}
    \item \textbf{An agent-native embodied evaluation protocol.}
    We introduce a minimal interface that leaves observation selection, information processing, task decomposition, and native action composition to the agent.

    \item \textbf{A capability-factorized manipulation benchmark.}
    Our framework supports \textit{200} tasks, with a 30-task primary suite
    covering perception, short-horizon execution, and long-horizon composition.

    \item \textbf{An empirical analysis of embodied agent capabilities.}
    We compare seven agents and show that the strongest agent's advantage concentrates on mechanism interaction, where sustained contact is strongly associated with success. Its remaining failures shift toward cross-stage interference and geometric errors.
\end{itemize}

\section{Related Work}
\subsection{Benchmarks for General-Purpose Agents}
Recent benchmarks increasingly evaluate language models as interactive agents. 
General-purpose suites such as AgentBench~\citep{liu2024agentbench}, AgentBoard~\citep{ma2024agentboard}, and GAIA~\citep{mialon2024gaia} test multi-step reasoning and tool use across diverse tasks.
WebArena~\citep{zhou2024webarena}, VisualWebArena~\citep{koh2024visualwebarena},
and WorkArena~\citep{drouin2024workarena}
focus on evaluating agents under diverse executable web environments.
OSWorld~\citep{xie2024osworld} and
AndroidWorld~\citep{rawles2025androidworld}
extend this setup to desktop and mobile computer use.
SWE-bench~\citep{jimenez2024swe} and AppWorld~\citep{trivedi2024appworld} test interaction with codebases and application APIs.
Together, these benchmarks emphasize multi-step interaction, feedback,
and execution-based outcomes.

We bring this principle to robot manipulation, where benchmark-side perception and control abstractions can themselves supply part of the capability being evaluated.
LIBERO-Agent therefore targets an \emph{agent-native} setting that leaves observation processing, task decomposition, and native action composition to the agent itself.

\subsection{Embodied Manipulation Benchmarks}

Robot manipulation benchmarks provide scalable and reproducible testbeds
for evaluating policies across diverse skills, scenes, and
horizons~\citep{yu2020meta,james2020rlbench,mees2022calvin,
liu2023libero,gu2023maniskill2,nasiriany2024robocasa}.
More recent benchmarks broaden capability coverage toward
language-conditioned reasoning~\citep{zhang2025vlabench},
memory-dependent tasks~\citep{lei2026robomemarena},
robust bimanual generalization~\citep{chen2025robotwin},
and compositional long-horizon execution~\citep{yang2026lilo}.
These benchmarks have substantially advanced robot-policy evaluation, but
typically assume a fixed observation interface.
In parallel, foundation-model approaches such as Code as
Policies~\citep{liang2023code} and VoxPoser~\citep{huang2023voxposer}
generate programs or spatial value maps executed through separate perception and control modules, separating high-level reasoning from
low-level physical execution.

Among existing works, EmbodiedBench~\citep{yang2025embodiedbench} and
CaP-X~\citep{fu2026cap} are most closely related to our setting.
EmbodiedBench evaluates multimodal language models on manipulation, but
reduces the burden of physical grounding and control through discretized
actions and additional object-level perceptual information.
CaP-X evaluates coding agents by synthesizing programs over predefined
robot perception and control primitives, and its seven-task core benchmark
primarily studies the effects of abstraction, interaction, and perceptual
grounding.
LIBERO-Agent instead evaluates agents through direct embodied interaction:
agents select and process observations themselves and compose bounded native action sequences that act on the simulator.
Our 30-task primary suite further factorizes evaluation into perception,
short-horizon execution, and long-horizon composition, enabling capability
diagnosis under the same interaction protocol.

\section{Benchmark}
\subsection{Agent--Simulation Interaction Protocol}
\label{sec:interaction-protocol}

We evaluate agents through iterative interaction with a simulated manipulation environment. At the beginning of each episode, an agent receives the task instruction, the available observation modalities, and a fixed interaction budget. The agent may inspect and process the available observations using its own tools, maintain information from previous interactions, and decide what to inspect, how to process it, and what actions to take. 

Interaction proceeds by alternating between agent decisions and environment execution. At each interaction step, the agent may submit a bounded sequence of robot actions. The simulation environment executes the sequence and returns updated observations. The agent can then use these new observations to continue working toward the task goal. No task-level manipulation primitives or privileged simulator state are exposed to the agent. Thus, the benchmark fixes the interaction interface and budgets, while leaving observation selection and action composition to the agent.

Episodes terminate when the agent explicitly finishes the task or exhausts the interaction or time budget. Task success is determined independently by the simulation environment. For tasks with process-level requirements, private evaluators additionally track relevant physical events, ordering constraints, and required prior events. 

\subsection{Simulation Implementation}
\label{sec:simulation-implementation}

We implement the interaction protocol on LIBERO~\citep{liu2023libero} through a minimal Model Context Protocol (MCP) interface with three tools. \texttt{start\_episode} initializes an episode, returns the task instruction and resource budgets, and publishes the initial observation. \texttt{osc\_sequence} executes a bounded sequence of robot actions and publishes the resulting observation, while \texttt{finish\_episode} terminates the episode and returns the final evaluation outcome.

The simulation is built on MuJoCo~\citep{todorov2012mujoco} with a 7-DoF Franka Emika Panda robot. Each \texttt{osc\_sequence} call accepts a bounded sequence of native \texttt{OSC\_POSE} actions. An action is represented by
$[\Delta x,\Delta y,\Delta z,r_x,r_y,r_z,g]\in[-1,1]^7$,
corresponding to relative end-effector translation, rotation, and gripper control. Translation and rotation components are scaled by 0.05~m and 0.5~rad, respectively, and the gripper command uses $g=-1$ for opening and $g=+1$ for closing. Submitted actions are executed by the environment's native \texttt{OSC\_POSE} controller.

By default, observations available to agents include third-person and wrist-view RGB images, metric depth, camera calibration, and robot state such as joint and end-effector states, velocities, commanded torques, and force/torque measurements. These observations are published to an episode-local workspace, where agents may inspect files, perform geometric computation, and store information across interactions. Each agent rollout is subject to fixed action-submission and wall-clock budgets. A separate evaluation process maintains private task checkers and event traces. 

Leveraging LIBERO's compatibility with related manipulation benchmarks, we integrate tasks and assets from LiLo-VLA~\citep{yang2026lilo}, RoboMemArena~\citep{lei2026robomemarena}, and VLABench~\citep{zhang2025vlabench}, resulting in a 200-task pool under a shared interaction interface.

\subsection{Task Design}
\label{sec:task-design}

From the 200-task pool supported by the simulation runtime, we select 30 representative tasks as the primary evaluation suite. We organize them into three task groups: \textbf{Perception-Focused Manipulation}, \textbf{Short-Horizon Manipulation}, and \textbf{Long-Horizon Manipulation}, with ten tasks in each group. Together, they separate three questions: can the agent identify the right target, execute the intended physical interaction, and compose these abilities across multiple stages? Short- and long-horizon tasks are further divided into easy and hard subsets based on difficulty.

\paragraph{Perception-Focused Manipulation.}
These tasks simplify physical interaction so that the main challenge lies in identifying task-relevant information. The ten tasks cover object pose, spatial relations, color, texture, instance markings, shape, semantic category, size, reflectance, and physical properties. Representative examples include identifying an upright marker, an object between two others, a different object, and the heaviest among visually identical objects. Perception is evaluated through embodied execution rather than question answering: agents must act on their judgments under a simple control requirement.

\paragraph{Short-Horizon Manipulation.}
These tasks specify both the relevant object and manipulation objective so that the main question is whether a correct target and goal can be turned into a reliable physical state change. We further divide them into easy and hard subsets according to execution difficulty: easy tasks allow relatively direct interactions and broader tolerances, whereas hard tasks require tighter geometric alignment, direction-sensitive motion, sustained contact, or more dexterous interaction. The easy subset consists of pick-and-place, button pressing, opening and closing a drawer, and pouring. The hard subset contains direction-sensitive pushing, opening a microwave, operating a stove control, precise tube insertion, and wiping a spill.

\paragraph{Long-Horizon Manipulation.}
These tasks compose the perception and manipulation abilities above across multiple stages, while adding planning and task-state tracking. We divide them into easy and hard subsets according to cross-stage difficulty: easy tasks mainly sequence familiar subgoals with limited cross-stage dependence, whereas hard tasks require stronger goal inference, causal dependencies, persistent physical constraints, or mechanism reasoning. Representative easy tasks include repeated pouring or transferring multiple objects, while hard tasks include reconstructing a scene from a goal image and using a mechanism such as a rotary latch or seesaw to complete the task. These tasks test whether agents can maintain a coherent task state and satisfy constraints over extended execution.

\section{Experiments}
\subsection{Experimental Setup}
\label{sec:experimental-setup}

\paragraph{Agents and primary evaluation.}
We evaluate seven general-purpose agents:
GPT-6 Astra~\citep{openai2026astra}, GPT-5.6 Sol~\citep{openai2026sol}, Claude Opus 5~\citep{anthropic2026opus5}, Claude Fable 5.1~\citep{anthropic2026fable51},
DeepSeek-V4.1-Flash~\citep{xu2026deepseek}, Kimi K3~\citep{team2026kimi}, and Qwen3.8-Max~\citep{alibabacloud2026qwen38max}.
Each evaluated agent is a model--harness pair, with reasoning effort set to high where supported.
Our primary evaluation uses the 30-task suite in Section~\ref{sec:task-design}
with the full observation profile and no demonstrations.
We evaluate each task with three rollouts using the same task-specific initial state and simulator seed.
Each rollout has a 1,800-second wall-clock limit and a fixed
action-submission budget.
A single \texttt{osc\_sequence} submission may contain between 1 and 50
native \texttt{OSC\_POSE} actions.
Third-person and wrist-view RGB and depth observations are rendered at
$256\times256$ resolution.

\paragraph{Metrics.}
We report \emph{stable} performance across three rollouts.
For perception and short-horizon tasks, let $y_{i,r}\in\{0,1\}$ denote full-task success for task $i$ in rollout $r$.
For a long-horizon task $i$ with $K_i$ stages, let $c_{i,r}$ denote the number of completed stages in rollout $r$.
Stable success rate and stable stage completion over a task set $\mathcal{T}$ are
\begin{equation}
\mathrm{SR}_{\mathrm{stable}}(\mathcal{T})
=\frac{100}{|\mathcal{T}|}\sum_{i\in\mathcal{T}}\prod_{r=1}^{3}y_{i,r},
\qquad
\mathrm{SC}_{\mathrm{stable}}(\mathcal{T})
=\frac{100}{|\mathcal{T}|}\sum_{i\in\mathcal{T}}
\frac{\min_{r\in\{1,2,3\}}c_{i,r}}{K_i},
\label{eq:stable-metrics}
\end{equation}
respectively.
Perception judgment accuracy (Judg. Acc.) follows the same stable criterion using correct target identification instead of full-task success.
Thus, $\mathrm{SR}_{\mathrm{stable}}$ requires full success in all three rollouts, while $\mathrm{SC}_{\mathrm{stable}}$ preserves partial progress while requiring it to be reproducible.
Tasks are equally weighted within each subset.
The 100-point overall Performance Score is
\begin{equation}
\mathrm{Score}
=5J+10P+5M_{\mathrm{easy}}+10M_{\mathrm{hard}}
+20L_{\mathrm{easy}}+50L_{\mathrm{hard}},
\label{eq:performance-score}
\end{equation}
where $J$ and $P$ denote stable perception judgment and task success, $M$ denotes stable short-horizon success, and $L$ denotes stable long-horizon stage completion, with all components normalized to $[0,1]$.
The weighting emphasizes harder and longer-horizon tasks.
Resource usage is averaged over all 90 primary episodes per agent.

\subsection{Main Results}
\label{sec:main-results}

\begin{table}[H]
\centering
\caption{Stable performance on the primary 30-task suite. Metrics follow the stable criterion in Section~\ref{sec:experimental-setup}. Judg. Acc.: target-identification accuracy; SR: success rate; SC: stage completion; M: million tokens. Score denotes the 100-point Performance Score. Bold marks the best task-performance result in each column.}
\label{tab:main-results}
\setlength{\tabcolsep}{4pt}
\renewcommand{\arraystretch}{1.00}
\resizebox{1.00\textwidth}{!}{%
\begin{tabular}{lccccccccc}
\toprule
Agent & \multicolumn{2}{c}{Perception} & \multicolumn{2}{c}{Short-Horizon} & \multicolumn{2}{c}{Long-Horizon} & Overall & \multicolumn{2}{c}{Resource Usage} \\
\cmidrule(lr){2-3}\cmidrule(lr){4-5}\cmidrule(lr){6-7}\cmidrule(lr){8-8}\cmidrule(lr){9-10}
& Judg. Acc. & Task SR & Easy SR & Hard SR & Easy SC & Hard SC & Score & Avg. Time & Avg. Tokens \\
\midrule
GPT-6 Astra & \textbf{100.0\%} & \textbf{100.0\%} & \textbf{100.0\%} & \textbf{40.0\%} & \textbf{50.0\%} & \textbf{22.0\%} & \textbf{45.0} & 11.5 min & 1.38M \\
GPT-5.6 Sol & 90.0\% & 10.0\% & 40.0\% & 0.0\% & 16.7\% & 0.0\% & 10.8 & 22.0 min & 3.44M \\
Claude Opus 5 & \textbf{100.0\%} & 80.0\% & 60.0\% & 0.0\% & 16.7\% & 0.0\% & 19.3 & 22.0 min & 5.49M \\
Claude Fable 5.1 & 90.0\% & 60.0\% & 40.0\% & 0.0\% & 16.7\% & 0.0\% & 15.8 & 22.4 min & 3.01M \\
DeepSeek-V4.1-Flash & 90.0\% & 0.0\% & 0.0\% & 0.0\% & 0.0\% & 0.0\% & 4.5 & 21.2 min & 0.39M \\
Kimi K3 & 90.0\% & 10.0\% & 0.0\% & 0.0\% & 0.0\% & 0.0\% & 5.5 & 19.0 min & 3.22M \\
Qwen3.8-Max & 90.0\% & 30.0\% & 20.0\% & 0.0\% & 10.0\% & 0.0\% & 10.5 & 23.9 min & 4.80M \\
\bottomrule
\end{tabular}%
}
\end{table}

\paragraph{Correct identification does not ensure reliable physical execution.}
Table~\ref{tab:main-results} reveals a pronounced gap between correct target identification and successful execution.
GPT-5.6 Sol, DeepSeek V4.1 Flash, Kimi K3, and Qwen3.8-Max all achieve 90\% stable judgment accuracy, yet their perception-focused task success is only 10\%, 0\%, 10\%, and 30\%, respectively.
The same weakness carries over to short-horizon manipulation: their easy-task success rates are only 40\%, 0\%, 0\%, and 20\%, and none succeeds stably on any hard short-horizon task.
Thus, these agents often identify \emph{what} to manipulate but fail to turn that decision into a reliable physical state change.

The gap narrows but persists for Opus and Fable.
Claude Opus 5 achieves 100\% judgment accuracy but 80\% task success, while Claude Fable 5.1 reaches 90\% and 60\%, respectively. Their easy short-horizon success also drops to 60\% and 40\%, and both score 0\% on the hard subset.
Only GPT-6 Astra achieves 100\% on both perception-focused metrics and easy short-horizon manipulation.
Overall, target identification is therefore already strong for most agents, while reliable physical execution remains much harder.

\paragraph{Performance degrades with interaction difficulty and task horizon.}
Reliable execution on simpler tasks does not necessarily extend to harder physical interactions.
Astra falls from 100\% success on easy short-horizon tasks to 40\% on the hard subset, while every other agent drops to 0\%.
The drop on hard short-horizon tasks shows that tighter alignment, direction-sensitive motion, sustained contact, and mechanism interaction already pose major challenges before multi-stage composition is introduced.
Long-horizon tasks further widen this gap.
Astra retains 50.0\% stable stage completion on the easy subset and 22.0\% on the hard subset.
Sol, Opus, and Fable each reach only 16.7\% on easy long-horizon tasks, Qwen reaches 10.0\%, and DeepSeek and Kimi make no stable progress. On the hard subset, every non-Astra agent scores 0\%.
These results suggest that longer horizons compound the difficulty of manipulation by requiring reliable execution across successive stages.
Even Astra, which consistently solves the simpler tasks, struggles to sustain this reliability over longer sequences.

\paragraph{Overall performance varies widely and is not explained by greater resource use.}
Overall, Astra achieves the highest Performance Score at 45.0/100, followed by Opus at 19.3 and Fable at 15.8.
Sol and Qwen obtain similar scores of 10.8 and 10.5, respectively, while Kimi and DeepSeek score 5.5 and 4.5.
Astra's performance advantage is accompanied by shorter episodes and lower token consumption than most other agents.
It averages 11.5 minutes per episode, compared with 19.0--23.9 minutes for the other agents.
Its average token usage is only 1.38M, below most other agents.
Taken together, Astra's higher success rates and lower resource use suggest that it completes tasks more efficiently, with less need for prolonged interaction or additional tokens.

\subsection{Observation and Demonstration Context}
\label{sec:observation-context}

\begin{table}[H]
\centering
\setlength{\tabcolsep}{3pt}
\renewcommand{\arraystretch}{0.96}

% Captions
\begin{minipage}[t]{0.485\textwidth}
\vspace{0pt}
\caption{Observation ablation: stable short horizon success rates (\%). Level~1: RGB + joint angles + end-effector pose; Level~2: Level~1 + dynamic proprioception; Level~3: Level~2 + calibrated depth. Bold marks the best nonzero result per agent and column.\strut}
\label{tab:observations}
\end{minipage}\hfill%
\begin{minipage}[t]{0.485\textwidth}
\vspace{0pt}
\caption{Demonstration ablation: stable long horizon stage completion (\%). None: no demo; Video: successful execution (MP4); Traj.: actions from another successful episode of the same task. Bold marks the best nonzero result per agent and column.\strut}
\label{tab:context}
\end{minipage}\par\noindent%
% Table bodies
\begin{minipage}[t]{0.485\textwidth}
\vspace{0pt}
\centering
\resizebox{0.94\linewidth}{!}{%
\begin{tabular*}{\linewidth}{@{\extracolsep{\fill}}lcccc@{}}
\toprule
Agent & Observation & \makebox[2.8em]{Easy} & Hard & Overall \\
\midrule
Astra & Level~1 & 40.0 & 20.0 & 30.0 \\
 & Level~2 & 80.0 & 20.0 & 50.0 \\
 & Level~3 & \textbf{100.0} & \textbf{40.0} & \textbf{70.0} \\
\addlinespace[1.5pt]
Opus & Level~1 & 20.0 & 0.0 & 10.0 \\
 & Level~2 & 20.0 & 0.0 & 10.0 \\
 & Level~3 & \textbf{60.0} & 0.0 & \textbf{30.0} \\
\addlinespace[1.5pt]
Fable & Level~1 & 0.0 & 0.0 & 0.0 \\
 & Level~2 & 0.0 & 0.0 & 0.0 \\
 & Level~3 & \textbf{40.0} & 0.0 & \textbf{20.0} \\
\bottomrule
\end{tabular*}%
}
\end{minipage}\hfill%
\begin{minipage}[t]{0.485\textwidth}
\vspace{0pt}
\centering
\resizebox{0.94\linewidth}{!}{%
\begin{tabular*}{\linewidth}{@{\extracolsep{\fill}}llccc@{}}
\toprule
Agent & Context & \makebox[2.8em]{Easy} & Hard & Overall \\
\midrule
Astra & None & 50.0 & 22.0 & 36.0 \\
 & Video & \textbf{93.3} & \textbf{39.0} & \textbf{66.2} \\
 & Video + Traj. & 90.0 & 26.0 & 58.0 \\
\addlinespace[1.5pt]
Opus & None & 16.7 & 0.0 & 8.3 \\
 & Video & 11.7 & 0.0 & 5.8 \\
 & Video + Traj. & \textbf{35.0} & 0.0 & \textbf{17.5} \\
\addlinespace[1.5pt]
Fable & None & 16.7 & 0.0 & 8.3 \\
 & Video & 30.0 & \textbf{5.0} & 17.5 \\
 & Video + Traj. & \textbf{30.8} & \textbf{5.0} & \textbf{17.9} \\
\bottomrule
\end{tabular*}%
}
\end{minipage}
\end{table}

\paragraph{Richer observations improve manipulation performance.}
To assess how different observation modalities affect manipulation success, we evaluate Astra, Opus, and Fable on the ten short-horizon tasks.
We compare three cumulative observation profiles.
Level~1 (L1) provides RGB observations, joint angles, and end-effector pose; Level~2 (L2) adds dynamic proprioception, including joint torques, end-effector velocity, and force/torque measurements; and Level~3 (L3) further adds metric depth and camera intrinsics and extrinsics.
The evaluation follows the protocol in Section~\ref{sec:experimental-setup}, with high reasoning effort and no demonstrations.

Table~\ref{tab:observations} shows that richer observations improve overall success, but the gains are mostly concentrated on easier tasks.
Adding dynamic proprioception raises Astra's overall success from 30\% to 50\%, driven by an increase from 40\% to 80\% on easy tasks, while Opus and Fable remain at 10\% and 0\%, respectively.
Adding calibrated depth then improves overall success by 20 percentage points for each agent, bringing Astra, Opus, and Fable to 70\%, 30\%, and 20\%, respectively.
For Opus and Fable, these gains occur entirely on easy tasks, where success rises from 20\% to 60\% and from 0\% to 40\%.
Astra benefits on both subsets, improving from 80\% to 100\% on easy tasks and from 20\% to 40\% on hard tasks.
We hypothesize that dynamic proprioception helps agents assess the robot's response to executed actions, while calibrated depth reduces uncertainty about the spatial relationship between the gripper and target objects.
These signals may therefore support more accurate actions and corrections during manipulation.

\paragraph{Demonstration gains depend on the agent and format.}
To examine how demonstration availability and format affect long-horizon execution, we evaluate Astra, Opus, and Fable on the ten long-horizon tasks under three conditions.
\textbf{None} uses the same no-demonstration setting as the primary evaluation.
\textbf{Video} additionally provides an MP4 video of successful task execution.
\textbf{Video + Traj.} further provides a successful action trajectory recorded in a different episode of the same task.
Direct replay of this trajectory does not solve the evaluation episode, but the trajectory provides action-level cues about task dynamics.
All conditions use full observations and high reasoning effort.
We report stable stage completion for the easy and hard subsets and across all ten tasks.

Table~\ref{tab:context} shows that video demonstrations improve long-horizon execution for Astra and Fable, whose overall stable stage completion rises from 36.0\% to 66.2\% and from 8.3\% to 17.5\%, respectively.
Opus, however, shows no clear benefit from video alone.
For Astra and Fable, video demonstrations yield larger gains on easy tasks than on hard tasks.
We hypothesize that videos clarify subgoal ordering and intermediate states, helping agents plan and monitor execution.
On harder tasks, however, difficult physical interaction may remain the dominant bottleneck even with this guidance, limiting further gains from video demonstrations.

Adding action trajectories does not consistently produce further gains over video alone.
Opus's overall stable stage completion increases from 5.8\% to 17.5\%.
Astra instead declines from 66.2\% to 58.0\%, mainly on hard tasks.
Fable gains little beyond video alone.
Since each reference trajectory comes from a different episode of the same task, we hypothesize that it provides useful low-level control cues but may also encourage actions that do not match the current episode.
Overall, all three agents improve over the no-demonstration baseline under at least one demonstration condition, but additional action-level information does not reliably yield further improvements.
\FloatBarrier

\section{Diagnosing the Strongest Agent}
\label{sec:astra_analysis}
\subsection{Localizing Astra's Long-Horizon Advantage}
\label{sec:astra_stage_analysis}

To understand where Astra's advantage comes from, we decompose the five
hard long-horizon tasks into 20 physical stages and manually annotate
whether each stage is reached, whether it succeeds, and its failure mode.
We compare Astra with the other three highest-scoring agents under the same high-reasoning, no-demonstration setting.

\begin{figure}[H]
    \centering
    \includegraphics[
        width=\linewidth,
        trim=0 0.5mm 0 2.5mm,
        clip
    ]{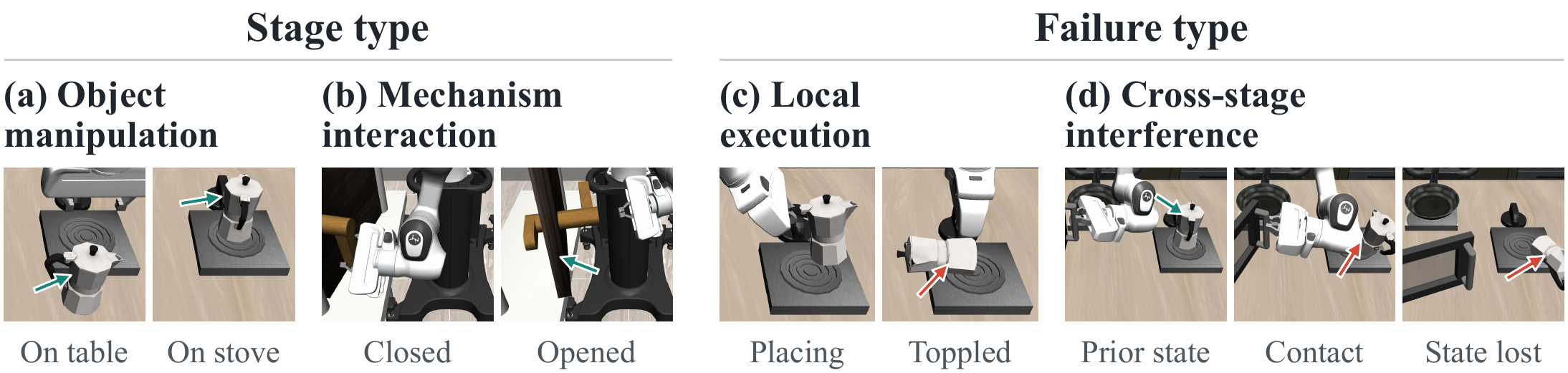}
    \caption{
    \textbf{Stage and failure taxonomy.}
    \textbf{(a--b)} Stage types: Object Manipulation and Mechanism Interaction;
    \textbf{(c--d)} failure types: Local Execution  and Cross-stage Interference.
    }
    \label{fig:stage_failure_taxonomy}
\end{figure}

We further categorize stages by their physical interaction and failures by where they arise, as illustrated in Figure~\ref{fig:stage_failure_taxonomy}.
\textit{Object Manipulation} moves or repositions a movable object, whereas
\textit{Mechanism Interaction} changes the state of an articulated mechanism.
For failed stages, \textit{Local Execution} denotes failure of the current physical operation itself, while
\textit{Cross-stage Interference} occurs when actions in one stage disturb, occlude, or undo a physical state established by another stage.
To isolate local physical competence, execution success rates are computed after excluding failures attributed to cross-stage interference.

\begin{table}[H]
\centering
\caption{
Stage-level diagnosis on the hard long-horizon tasks.
\textit{Reached} and \textit{Success} are computed over all stage
opportunities.
Object Manipulation and Mechanism Interaction report execution success
rates after excluding failures attributed to cross-stage
interference.
Failure modes are percentages of reached-but-failed stages and may overlap.
}
\label{tab:stage_diagnosis}
\small
\setlength{\tabcolsep}{6pt}
\resizebox{\textwidth}{!}{
\begin{tabular}{lcccccc}
\toprule
& \multicolumn{2}{c}{Stage Progress (\%)}
& \multicolumn{2}{c}{Execution SR (\%)}
& \multicolumn{2}{c}{Failure Mode (\%)} \\
\cmidrule(lr){2-3}
\cmidrule(lr){4-5}
\cmidrule(lr){6-7}
Agent
& Reached
& Success
& Object Manipulation
& Mechanism Interaction
& Local Execution
& Cross-stage Interference \\
\midrule
GPT-6 Astra
& \textbf{65.0}
& \textbf{36.7}
& \textbf{66.7}
& \textbf{94.1}
& \textbf{23.5}
& 52.9 \\

GPT-5.6 Sol
& 43.3
& 13.3
& \textbf{66.7}
& 22.2
& 61.1
& 27.8 \\

Claude Opus 5
& 41.7
& 5.0
& 28.6
& 7.1
& 81.8
& 0.0 \\

Claude Fable 5.1
& 26.7
& 6.7
& 20.0
& 37.5
& 83.3
& 0.0 \\
\bottomrule
\end{tabular}
}
\end{table}

Table~\ref{tab:stage_diagnosis} shows that Astra's advantage appears at
both task progression and physical execution.
It reaches 65.0\% of all stages and successfully completes 36.7\%,
substantially exceeding the other agents on both measures.
The largest execution gap appears in mechanism interaction:
Astra reaches a 94.1\% execution success rate, while the other agents
range from 7.1\% to 37.5\%.
By contrast, Astra and Sol attain the same 66.7\% rate on object
manipulation.
The failure annotations reinforce this distinction.
Local execution accounts for 61.1--83.3\% of reached-stage failures for
the weaker agents, but only 23.5\% for Astra.
These results therefore localize Astra's long-horizon advantage primarily
to stronger mechanism-level physical execution.

\subsection{Sustained Contact During Mechanism Interaction}
\label{sec:astra_contact_analysis}

We next examine what Astra does differently during mechanism interaction.
We focus on opening operations in which the gripper must remain physically
engaged with a handle while the articulated mechanism moves.
We quantify this behavior using the \emph{sustained-contact ratio}:
\begin{equation}
    N_{\mathrm{sustained}}
    = \sum_{t=1}^{N_{\mathrm{exec}}}
    \mathbf{1}[\ell_t \ge 5],
    \qquad
    \mathrm{SCR}
    = \frac{N_{\mathrm{sustained}}}{N_{\mathrm{exec}}}\times 100\%,
\end{equation}
where $\ell_t$ is the length of the gripper--handle contact run containing
control step $t$, with $\ell_t=0$ when no contact occurs.

\begin{figure}[H]
    \centering
    \includegraphics[
        width=\linewidth,
        trim=0 1.5mm 0 5mm,
        clip
    ]{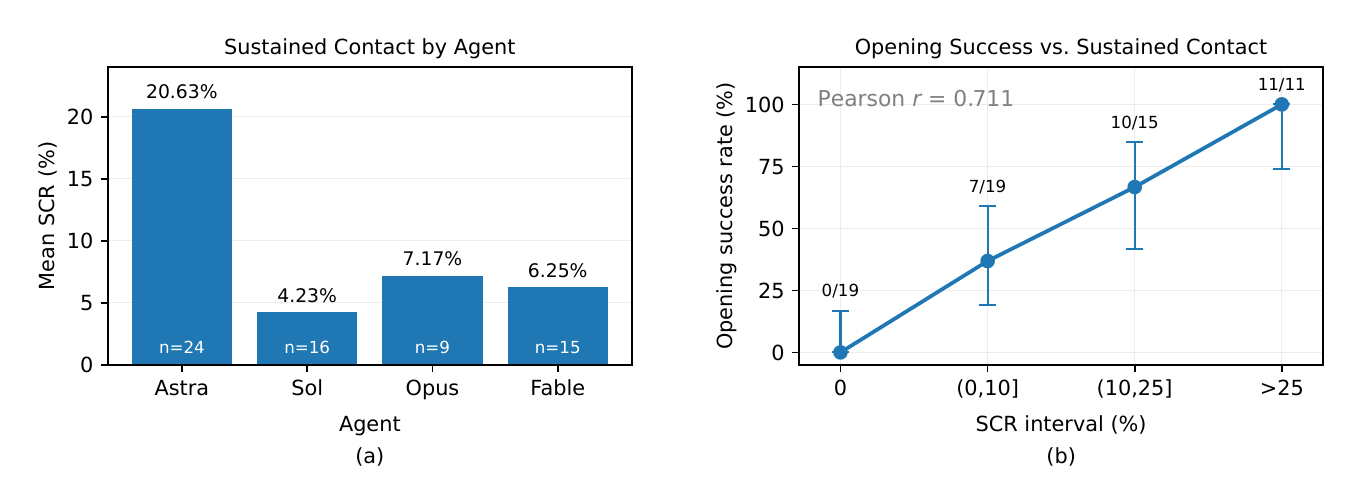}
    \caption{
    \textbf{Sustained contact during mechanism interaction.}
    \textbf{(a)} Mean sustained-contact ratio (SCR) across agents for
    entered opening operations.
    \textbf{(b)} Local opening success across SCR intervals. Error bars in (b) show 95\% Wilson confidence intervals. SCR intervals are used only for visualization. Pearson correlation is computed over the original 64 opening operations.
    }
    \label{fig:sustained_contact}
\end{figure}

Figure~\ref{fig:sustained_contact} shows a substantial difference between
Astra and the other agents.
Astra reaches a mean SCR of 20.63\%, whereas the other agents range from
4.23\% to 7.17\%.
The same measure also tracks local opening success.
Among operations with zero SCR, none of 19 succeed.
The success rate rises to 7/19 for SCR between 0 and 10\%, to 10/15 for
SCR between 10 and 25\%, and to 11/11 when SCR exceeds 25\%.
Using the original 64 opening operations rather than the visualization bins,
SCR has a Pearson correlation of $r=0.711$ with local opening success. Together, these results associate Astra's higher opening success with more
consistent sustained physical engagement.

\subsection{Remaining Failures: Cross-stage Interference and Geometry}
\label{sec:astra_remaining_failures}

The preceding analyses show that Astra has largely reduced the local
execution failures that dominate weaker agents, particularly during
mechanism interaction.
Its remaining errors therefore show a different failure pattern:
successfully completing one physical interaction does not guarantee that
the resulting scene remains suitable for subsequent stages.

\begin{figure}[H]
    \centering
    \includegraphics[
        width=\linewidth,
        trim=0 0.5mm 0 2.5mm,
        clip
    ]{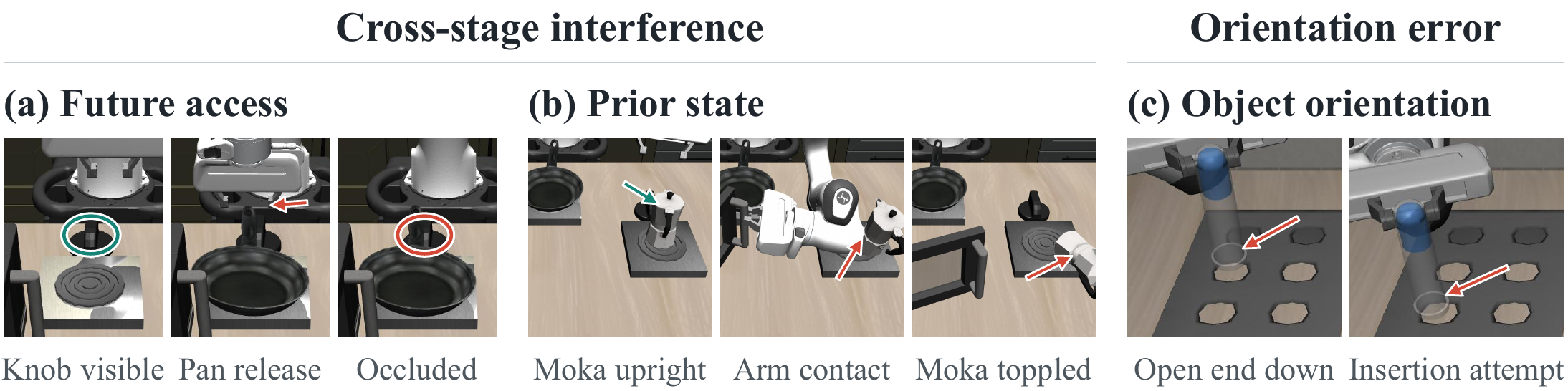}
    \caption{
    \textbf{Residual failure modes of Astra.}
    (a) A placement blocks access required by a later stage.
    (b) A subsequent action disturbs previously established state.
    (c) The intended operation is executed with an invalid object orientation.
    }
    \label{fig:astra_failure_cases}
\end{figure}

Figure~\ref{fig:astra_failure_cases} illustrates two recurring forms of
failure.
First, actions interfere across stages.
A reasonable placement can occlude or block a later interaction, while a
subsequent motion can disturb an object or mechanism whose state was
already correct.
Second, Astra can select the correct object and operation while realizing
it with invalid geometry.
For example, it may attempt tube insertion with the tube in the wrong
orientation.

These failures indicate that stronger local manipulation alone is not
sufficient for reliable long-horizon control.
Astra can execute individual physical interactions effectively, yet still
struggles to preserve task-relevant state across stages and to realize
actions with task-valid geometry.

\FloatBarrier

\section{Conclusion}
\label{sec:conclusion}
We introduced LIBERO-Agent, a benchmark for evaluating general-purpose agents through autonomous observation selection and native robot control. Its perception-focused, short-horizon, and long-horizon tasks provide complementary tests of embodied competence. GPT-6 Astra achieves the highest Performance Score of 45.0/100 among seven evaluated agents. It excels at target identification and easy short-horizon manipulation, but remains less reliable on harder short-horizon and long-horizon tasks. More broadly, our benchmark reveals a common limitation across agents: correct target identification does not consistently translate into reliable physical execution. Our ablations show that richer observations improve manipulation, while demonstration benefits vary by agent and format. Further analysis shows that Astra's advantage is strongest in mechanism interaction and is associated with sustained physical contact. As local execution improves, its remaining failures shift toward cross-stage interference and geometric errors, highlighting anticipation of downstream consequences and preservation of previously achieved states as key remaining challenges.

\clearpage
\bibliographystyle{plainnat}
\bibliography{main}

\clearpage
\appendix
\section{Agent Harnesses}
\label{app:agent-harnesses}

Table~\ref{tab:app-harness-versions} lists the agent harness version used for each evaluated model. Versions are taken from the archived
run manifests and session metadata.

\begin{table}[htbp]
\centering
\caption{Agent harnesses and versions used in our experiments.}
\label{tab:app-harness-versions}
\small
\setlength{\tabcolsep}{8pt}
\renewcommand{\arraystretch}{1.15}
\begin{tabular}{@{}lll@{}}
\toprule
Model & Agent harness & Version \\
\midrule
GPT-6 Astra        & Codex           & 0.154.0 \\
GPT-5.6 Sol        & Codex           & 0.154.0 \\
\addlinespace
Claude Opus 5      & Claude Code     & 2.1.139 \\
Claude Fable 5.1   & Claude Code     & 2.1.139 \\
\addlinespace
DeepSeek V4.1 Flash & DeepSeek Harness & 0.1.5-rc.2 \\
Kimi K3           & Kimi Code       & 0.42.0 \\
Qwen3.8-Max        & Qwen Code       & 0.24.0 \\
\bottomrule
\end{tabular}
\end{table}

\section{Task Suite and Evaluation Details}
\label{app:tasks-evaluation}

\subsection{Task construction and primary evaluation suite}
\label{app:task-catalog}

Our primary evaluation comprises 30 tasks: ten perception-focused tasks,
ten short-horizon manipulation tasks, and ten long-horizon tasks.
Tables~\ref{tab:app-perception-tasks}--\ref{tab:app-long-tasks} enumerate
their instructions, physical requirements, and scoring units.
The task collection combines LIBERO scenes and predicates~\citep{liu2023libero}, RoboMemArena task sequences~\citep{lei2026robomemarena}, LiLo-VLA task scenes~\citep{yang2026lilo}, VLABench assets and
mechanisms \citep{zhang2025vlabench}, and adapted manipulation mechanisms
in the shared MuJoCo/robosuite runtime \citep{zhu2020robosuite}.
Task adaptation specifies the scene, initial state, public instruction,
and evaluator together. The same observation and control interface is
used across these sources.

The perception suite varies the information needed to select a target.
Most tasks pair identification with pick-and-place or pressing a paired
button. T08 instead requires physical probing: its three mugs have the
same appearance, and the agent must lift them to distinguish their
physical properties. Target identification and completion of the requested
physical operation are scored separately.

The short-horizon split groups T11, T13--T15, and T20 as easy tasks,
and T12 and T16--T19 as hard tasks. These groups distinguish elementary
grasping and actuation from more constrained pushing, articulated motion,
orientation control, insertion, and contact coverage.
The long-horizon easy subset comprises T23, T24, T26, T27, and T29;
the hard subset comprises T21, T22, T25, T28, and T30.
Its harder tasks combine goal interpretation, temporary-buffer planning,
persistent contact constraints, or coupled mechanism interactions.
Each difficulty subset contains five tasks and receives equal task
weighting internally.

\begin{table}[htbp]
\centering
\caption{Perception-focused tasks. The target-selection requirement is
distinct from the complete physical outcome. The identifiers T01--T30
are shared across the task catalogs and execution examples.}
\label{tab:app-perception-tasks}
\small
\setlength{\tabcolsep}{4pt}
\renewcommand{\arraystretch}{1.15}
\begin{tabular}{@{}p{0.065\linewidth}p{0.22\linewidth}p{0.64\linewidth}@{}}
\toprule
ID & Primary information & Required physical outcome \\
\midrule
T01 & Object pose & Place the only upright marker in the collection bin. \\
T02 & Spatial relation & Place the object between the other two objects in the basket. \\
T03 & Color & Place the red object in the collection bin. \\
T04 & Texture & Place the striped object in the collection bin. \\
T05 & Instance marking & Place the object with exactly one square mark in the collection bin. \\
T06 & Shape & Press the button paired with the small ball, among ball, cylinder, and cube candidates. \\
T07 & Object category & Place the cola in the collection bin. \\
T08 & Physical property & Physically sample all three visually identical mugs and place the heaviest in the green bin. \\
T09 & Reflectance & Press the button paired with the most reflective object. \\
T10 & Size & Press the button paired with the largest cylinder. \\
\bottomrule
\end{tabular}
\end{table}

\begin{table}[htbp]
\centering
\caption{Short-horizon manipulation tasks. Easy and hard refer to the
evaluation split. A short-horizon task can contain several physical
events, but its reported task success requires the complete operation.}
\label{tab:app-short-tasks}
\small
\setlength{\tabcolsep}{4pt}
\renewcommand{\arraystretch}{1.15}
\begin{tabular}{@{}p{0.06\linewidth}p{0.075\linewidth}p{0.22\linewidth}p{0.54\linewidth}@{}}
\toprule
ID & Split & Task & Success requirement \\
\midrule
T11 & Easy & Pick and place & Alphabet soup is in the basket. \\
T12 & Hard & Push plate & The plate reaches the specified region in front of the stove. \\
T13 & Easy & Press button & The red button is physically actuated. \\
T14 & Easy & Close drawer & The cabinet's top drawer reaches its closed state. \\
T15 & Easy & Open drawer & The cabinet's top drawer reaches its open state. \\
T16 & Hard & Open microwave & The microwave door reaches its open state. \\
T17 & Hard & Turn on stove & The stove knob reaches the activation condition. \\
T18 & Hard & Insert tube & The blue chemistry tube satisfies the target slot's position, vertical alignment, and motion tolerances. \\
T19 & Hard & Wipe spill & Grasp the sponge, cover the spill through physical wiping contact, and release the sponge on the blue rest area. \\
T20 & Easy & Pour into mug & Lift the wine bottle and complete a physical pour event over the mug. \\
\bottomrule
\end{tabular}
\end{table}

\begin{table}[htbp]
\centering
\caption{Long-horizon task requirements and scoring units. $K$ denotes
the number of scored stages or goals. Cooking goals are evaluated at
the terminal state without an imposed order.}
\label{tab:app-long-tasks}
\small
\setlength{\tabcolsep}{4pt}
\renewcommand{\arraystretch}{1.15}
\begin{tabular}{@{}p{0.055\linewidth}p{0.07\linewidth}p{0.045\linewidth}p{0.725\linewidth}@{}}
\toprule
ID & Split & $K$ & Goals or stages \\
\midrule
T21 & Hard & 5 & Prepare the cooking area from text: moka pot on its stove, that stove on, frying pan on its stove, that stove on, and microwave open. \\
T22 & Hard & 5 & Match the cooking goal image, using the same five physical goals as T21. \\
T23 & Easy & 8 & Open and close the top, middle, and bottom drawers in order; reopen the occupied drawer; place butter inside it. Final closure is optional. \\
T24 & Easy & 4 & Lift the tomato-sauce bottle, complete two distinct pours over the frying pan, and place the bottle in the bowl drainer. \\
T25 & Hard & 4 & Move the left bowl to the empty plate, the middle bowl to the left plate, the right bowl to the middle plate, and the buffered bowl to the right plate. No bowl may touch the tabletop. \\
T26 & Easy & 3 & Transfer tomato sauce, milk, and orange juice from the first cabinet to the second cabinet. \\
T27 & Easy & 4 & Open the middle drawer, place cookies inside, place chocolate inside, and close the drawer. \\
T28 & Hard & 2 & Release the rotating latch and open the door. \\
T29 & Easy & 3 & Lift the hammer, establish hammer--nail contact, and drive the nail into the block. \\
T30 & Hard & 4 & Place the first and second counterweights, produce a physical lift through the seesaw, and place the exposed target in the basket. \\
\bottomrule
\end{tabular}
\end{table}

\subsection{Task-specific success semantics}
\label{app:success-semantics}

The evaluators combine terminal predicates with physical event histories
according to the task instruction. Object placement and ordinary
articulation tasks use the corresponding containment, support, or joint
predicates. Tasks involving counting, support constraints, or causal
dependencies additionally record intermediate events. Full success and
partial stage completion are stored separately.

\paragraph{Paired textual and visual goals.}
T21 and T22 share a scene definition, an initial-state bank, and five
terminal predicates. T21 communicates the desired arrangement in text,
whereas T22 supplies a goal image. Both accept any execution order that
satisfies the five goals at termination. The evaluator records changes in
the set of satisfied goals, including regressions when later actions
disturb an earlier placement or mechanism state. A goal that was satisfied
earlier but is false at termination does not contribute to the automatic
terminal-goal count.

\paragraph{Ordered search and placement.}
T23 checks the drawer inspection sequence before the final butter
placement. The occupied target drawer is determined from the initial
scene. Its eight required stages exclude an optional final closure.
T27 instead explicitly requires closure after the two placements, so
closure contributes to its four-stage score.
The evaluator records ordered progress and checks the required final
object arrangement.

\paragraph{Repeated events and persistent constraints.}
For T24, two pours must be distinct physical events; the pour counter
tracks bottle elevation, tilt, and placement relative to the receptacle,
and rejects a detected third pour. T25 uses the empty plate as a temporary
buffer and records the four ordered transfers. Bowl--table contact is
monitored throughout execution, and the final buffer must be empty.
T30 requires released counterweights to contact the movable board before
the mechanism lifts the target. At completion, the target must be in the
basket while both counterweights remain supported at the far end.

\paragraph{Active sensing and contact-dependent operations.}
T08 uses three visually identical mugs with masses of 0.18, 0.36, and
0.72\,kg. Its automatic evaluator requires grasp-and-lift evidence for
each mug, followed by placement of the heaviest mug in the green bin,
released and clear of the gripper. Mass labels remain evaluator-private.
T19 requires the grasped sponge to cover seven spill markers during
physical table contact before it is released on the rest area.
T29 requires hammer--nail contact before accepting a nail displacement
of at least 0.075\,m.

\paragraph{Insertion geometry.}
The automatic tube checker in T18 tests lateral error of at most
0.018\,m, a tube-root height relative to the rack root in
$[0.032,0.085]$\,m, absolute tube-axis alignment with the vertical of at
least 0.97, linear speed of at most 0.08\,m/s, and angular speed of at
most 0.60\,rad/s. Absolute axis alignment measures verticality and is
insensitive to end-for-end reversal.

\subsection{Episode conditions and outcome adjudication}
\label{app:episode-conditions}

The primary comparison evaluates GPT-6 Astra, GPT-5.6 Sol, Claude Opus~5,
Claude Fable~5.1, DeepSeek V4.1 Flash, Kimi K3, and Qwen3.8-Max.
Each configuration contributes three executions of every primary task,
for 90 episodes per agent. The simulator seed is fixed at 100, and the
task-specific initial state is held fixed across the three executions.
These repetitions measure variation in agent execution under the same
scene conditions.

Primary runs use high reasoning effort where supported, full observations,
and no demonstrations. Each episode has a
1,800-second evaluator wall-clock limit. The recorded prompts specify a
ceiling of 1,000,000 accepted OSC sequence submissions, with 1--50 native
actions per submission; the wall-clock limit is the operative execution
budget. A native action advances one 20\,Hz control interval.
The agent can write local programs and inspect selected files in its
episode workspace. Benchmark observations and evaluator-private records
are kept separate in the data interface.

\clearpage
\section{Task Prompts and Demonstration Context}
\label{app:prompts-context}

\subsection{Shared episode prompt}

Each episode provides the task instruction followed by a shared execution
prompt. Figure~\ref{fig:app-episode-prompt} reproduces the no-demonstration
prompt from the successful T23 execution used in
Appendix~\ref{app:successful-programs}. Line breaks are added for layout.
The reference to reset attempts is retained from the recorded template;
the evaluated configuration exposes no reset operation.
Task-specific strategy and control programs are supplied by the agent
during execution.

\begin{figure}[htbp]
\centering
\begin{minipage}{0.96\linewidth}
\hrule
\vspace{6pt}
\textbf{Recorded task and shared execution prompt}
\small
\begin{verbatim}
Open and close all drawers in order to check. Put butter into the
drawer that already contains an object.

A LIBERO episode has been prepared for you.

This query has a 1800-second evaluator wall-clock budget shared by
all reset attempts. Exceeding it ends the query as an official
task failure.

1. Call the `start_episode` robot tool exactly once to begin and
   receive the initial observation and `max_agent_steps` budget.
2. Control the robot with the `osc_sequence` robot tool. Its
   `actions` argument is an array of 1 to 50 normalized 7D OSC_POSE
   micro actions in `[dx, dy, dz, rx, ry, rz, gripper]` order.
   Every component must be within [-1, 1]. Each vector executes
   one LIBERO policy interval; translation 1.0 corresponds to
   0.05 m, rotation 1.0 to a 0.5 rad rotation-vector component,
   gripper -1 opens, and +1 closes. One sequence call counts as
   one Agent action, with at most 1000000 accepted calls.
3. Wait for each action to complete, then inspect
   `benchmark_inputs/current_observation/observation.json` and any
   referenced files before issuing another command.
4. When you have completed the task, call the `finish_episode`
   robot tool exactly once. Only finish reports official task
   success.
\end{verbatim}
\hrule
\end{minipage}
\caption{Shared episode instructions, shown with the T23 task text.
The prompt describes the interaction contract and points to the current
observation. It supplies no drawer-opening or object-transfer program.}
\label{fig:app-episode-prompt}
\end{figure}

For demonstration-conditioned episodes, the following notice is inserted
after the task instruction:
\begin{quote}
\small
A demonstration from a separate episode of the same task is available at
\nolinkurl{benchmark_inputs/expert_demo/}. The current scene configuration
and object or goal poses may differ.
\end{quote}
Both demonstration conditions use this notice. The available files
determine whether the reference contains video alone or video together
with an action trajectory.

\subsection{Representative task instructions}

Task instructions state the desired operation and its public constraints.
The following examples preserve the recorded task wording:
\begin{quote}
\small
\textbf{T08, physical perception.}
Determine which of the three visually identical mugs is heaviest by
physically weighing them, then place the heaviest mug in the green bin.

\textbf{T18, precision manipulation.}
Pick up the blue chemistry tube and insert it vertically into the
rear-center empty slot of the tube rack.

\textbf{T25, constrained rearrangement.}
Cyclically rearrange the three black bowls from left to right in the head
camera. No bowl may touch the tabletop at any time.

\textbf{T28, coupled articulation.}
Open the door. It is secured by a rotating latch.

\textbf{T30, mechanism use.}
Lift the hidden object and place it in the basket.
\end{quote}
The weighing prompt requires physical sampling but does not identify the
heaviest mug. The rearrangement prompt communicates the support constraint
without naming the empty plate as a buffer. The seesaw prompt leaves the
mechanism-use strategy to the agent and, when available, its demonstration.

\subsection{Video and action-trajectory bundles}
\label{app:demonstration-bundles}

Demonstrations come from separate successful executions of the same task.
Their public representation is specified in
Table~\ref{tab:app-context-bundles}. The video contains synchronized head
and wrist RGB views displayed side by side at 20 frames per second.
Each view also has a contact sheet with at most 12 uniformly sampled
frames, retaining the sequence endpoints. These sheets provide a directly
viewable overview; the agent can inspect the video further using its own
tools.

\begin{table}[htbp]
\centering
\caption{Contents of the demonstration bundle. These columns describe
historical reference data; the live observation profile is held fixed
across the context comparison.}
\label{tab:app-context-bundles}
\small
\setlength{\tabcolsep}{5pt}
\renewcommand{\arraystretch}{1.15}
\begin{tabular}{@{}p{0.51\linewidth}ccc@{}}
\toprule
Reference content & None & Video & Video + Traj. \\
\midrule
Head/wrist RGB video & --- & Yes & Yes \\
RGB contact sheets & --- & Yes & Yes \\
Ordered native OSC action vectors & --- & --- & Yes \\
Reference depth, calibration, or robot-state stream & --- & --- & --- \\
Private object states or checker progress & --- & --- & --- \\
\bottomrule
\end{tabular}
\end{table}

Video + Traj. adds \nolinkurl{trajectory.jsonl}. Each row pairs a
zero-based \nolinkurl{action_index} with an \texttt{action} object.
The object contains a schema identifier and a
\nolinkurl{normalized_vector_7d} field. Its seven components follow the
online OSC convention: three translation components, three rotation-vector
components, and one gripper command. All values lie in $[-1,1]$, with one
action vector per policy interval. The trajectory contains issued actions;
it does not include a numerical stream of measured end-effector poses.

Bundle validation checks the task instruction, file inventory, action
dimensions and ranges, contiguous action indices, and consistency of the
declared frame and action counts. The agent receives a manifest, the RGB
video, two contact sheets, and, for Video + Traj., the action file.
Current scene observations remain distinct from historical demonstration
content, allowing the agent to adapt its execution to the observed scene.

\clearpage
\section{Programs Constructed During Successful Executions}
\label{app:successful-programs}

We examine two successful GPT-6 Astra episodes under high reasoning
effort, full observations, and no demonstrations: drawer inspection and
placement (T23), and physical weighing (T08).
The examples connect visible program construction and tool use to
recorded observations, executed actions, and final task outcomes.
Code excerpts preserve the recorded computations, with line wrapping and
omission of unrelated display operations. The programs were created
during these episodes rather than supplied as task-level primitives.

\subsection{Computing spatial coordinates from selected image locations}
\label{app:depth-program}

At the beginning of T23, Astra inspects the scene and samples selected
pixels from the head-camera depth array. It loads the camera intrinsics
and camera-to-robot-base transform, back-projects each selected pixel,
and prints its three-dimensional coordinate.
Figure~\ref{fig:app-depth-program} shows two returned surface points.
The agent subsequently issues an approach command with an end-effector
target at $(0.63,0.34,0.273)$\,m before moving closer.

The pixel locations are selected by the agent. Their back-projected
coordinates are measurements of visible surfaces, from which the agent
chooses an end-effector target with an approach offset. This trace shows
how publicly available depth and calibration enter local computation
before control. Object poses and grasp targets are not supplied by the
evaluator.

\begin{figure}[htbp]
\centering
\begin{minipage}{0.96\linewidth}
\hrule
\vspace{6pt}
\textbf{T23: agent-written depth back-projection}
\small
\begin{verbatim}
import json, numpy as np
p = 'benchmark_inputs/current_observation/'
o = json.load(open(p + 'observation.json'))
c = o['cameras']['head']
d = np.load(p + c['depth']['metric_file'])
K = np.array(c['intrinsic_matrix_3x3'])
T = np.array(c['matrix_T_robot_base_from_camera_opencv_4x4'])
for v in [124,128,132,140,144,148,152,156]:
    for u in [214,217,220]:
        a = T @ np.r_[
            np.linalg.inv(K) @ np.array([u,v,1]) * d[v,u], 1]
        print((u,v), a[:3].round(3))
\end{verbatim}
\hrule
\vspace{6pt}
\textbf{Selected tool output}
\begin{verbatim}
(217, 128) [0.642 0.428 0.272]
(220, 128) [0.643 0.442 0.273]
\end{verbatim}
\hrule
\end{minipage}
\caption{Numerical geometry constructed from public observations in a
successful drawer-search episode. The excerpt and output come from the
recorded tool call. Coordinates are expressed in the robot-base frame
and measured in metres. The agent selects the pixels and subsequent
approach targets.}
\label{fig:app-depth-program}
\end{figure}

\subsection{Constructing and reusing a pose-feedback helper}
\label{app:pose-program}

In the same T23 episode, Astra creates a local Python helper,
\texttt{robot\_pose.py}. Each invocation reads the current end-effector
pose, computes position and rotation errors relative to command-line
targets, clips the resulting normalized control values, and prints a
JSON action sequence. Figure~\ref{fig:app-pose-program} shows the helper's
action-generation branch and its subsequent connection to the MCP tool.
The rotation error uses the relative rotation
$R_{\mathrm{target}}R_{\mathrm{current}}^{-1}$ in rotation-vector form.

The helper repeats one computed action within a submitted batch.
Feedback is refreshed between batches, when the agent invokes the helper
again on the newly published observation. In the initial orientation
adjustment, six consecutive batches execute 60 control steps while
retaining the same target orientation. The measured angular error falls
from $121.88^\circ$ to $0.76^\circ$.
Table~\ref{tab:app-pose-chains} reports this and three subsequent
same-orientation correction sequences from the episode. Positions can
change while the orientation target is held fixed.

The evaluator records all eight required T23 stages as completed. The
episode uses 57 accepted control submissions and 895 native control
steps; the agent also performs the optional final drawer closure.
The first drawer opening is recorded at control step 205, and the
required butter placement at step 732. These events connect the
programming example to a complete successful task execution.

\begin{figure}[htbp]
\centering
\begin{minipage}{0.96\linewidth}
\hrule
\vspace{6pt}
\textbf{T23: action-generation branch of \texttt{robot\_pose.py}}
\small
\begin{verbatim}
import json, sys, numpy as np
from scipy.spatial.transform import Rotation as R
p = 'benchmark_inputs/current_observation/'
o = json.load(open(p + 'observation.json'))
s = o['state']
pose = s['eef_pose_robot_base_xyzw_7d']
if len(sys.argv) > 1:
    target = np.array([float(x) for x in sys.argv[1:4]])
    q = np.array([float(x) for x in sys.argv[4:8]])
    g = float(sys.argv[8])
    n = int(sys.argv[9])
    dp = target - pose[:3]
    dr = (R.from_quat(q) *
          R.from_quat(pose[3:]).inv()).as_rotvec()
    a = np.r_[np.clip(dp*5,-1,1), np.clip(dr*.8,-1,1), g]
    print(json.dumps({'actions': [a.tolist() for _ in range(n)]}))
\end{verbatim}
\hrule
\vspace{6pt}
\textbf{Recorded invocation and action submission}
\begin{verbatim}
const r = await tools.exec_command({
  cmd: "python /tmp/robot_pose.py .63 .34 .273 " +
       "0 .70710678 .70710678 0 -1 10",
  max_output_tokens: 3000
});
text(await tools.mcp__libero__osc_sequence(JSON.parse(r.output)));
\end{verbatim}
\hrule
\vspace{6pt}
\textbf{Execution receipt, selected fields}
\begin{verbatim}
"command_completed": true, "control_steps": 10,
"termination_reason": "sequence_executed",
"observation_id": "obs_000002"
\end{verbatim}
\hrule
\end{minipage}
\caption{Astra constructs a pose-feedback helper and passes its numerical
output to native OSC control. The helper is generated in the session.
Its action-generation branch is shown with whitespace expanded; the
read-only diagnostic branch is omitted. The shell-command string is
split for layout. A receipt establishes execution of the sequence;
task completion is checked separately.}
\label{fig:app-pose-program}
\end{figure}

\begin{table}[htbp]
\centering
\caption{Repeated orientation corrections in the successful T23 episode.
Each sequence uses fresh state before successive control batches and
retains a common target orientation. Angular error is measured from the
recorded end-effector quaternion to the agent-selected target.}
\label{tab:app-pose-chains}
\small
\begin{tabular}{@{}ccccc@{}}
\toprule
Sequence & Control batches & Control steps & Initial error & Final error \\
\midrule
1 & 6 & 60 & $121.88^\circ$ & $0.76^\circ$ \\
2 & 4 & 42 & $89.68^\circ$ & $2.26^\circ$ \\
3 & 5 & 61 & $87.99^\circ$ & $0.14^\circ$ \\
4 & 4 & 46 & $90.42^\circ$ & $0.61^\circ$ \\
\bottomrule
\end{tabular}
\end{table}

\subsection{Acquiring physical evidence through active probing}
\label{app:weighing-program}

T08 provides a complementary example in which the necessary evidence is
obtained through interaction. Astra successively grasps and lifts the
three visually identical mugs. For each mug it executes a batch of
20 upward commands followed by 20 commands with zero pose increments
and a closing gripper command. It then reads the current state and
proprioception and inspects an RGB image.
Figure~\ref{fig:app-weighing-program} shows this repeated action pattern
and the corresponding force readings.

The sensor-frame $z$ components after the three lift-and-hold batches
are $-8.64$, $-12.19$, and $-6.86$\,N. The middle mug produces the
largest load magnitude, and the agent subsequently places it in the
green bin. The private evaluator confirms that all three mugs were
physically sampled and that the heaviest was placed, released, and left
clear of the gripper. The episode completes all four required events
in 28 accepted control submissions and 614 native control steps.

These are raw end-effector sensor readings, which include the gripper
load and are expressed in the sensor frame. The trace demonstrates
repeated physical probing followed by successful target selection;
it does not contain an explicit calibrated mass-estimation program.
Here, zero pose increments create a hold interval for sensing within
the manipulation sequence.

\begin{figure}[htbp]
\centering
\begin{minipage}{0.96\linewidth}
\hrule
\vspace{6pt}
\textbf{T08: recorded lift-and-hold action pattern}
\small
\begin{verbatim}
text(await tools.mcp__libero__osc_sequence({actions: [
  ...Array.from({length:20}, () => [0,0,0.6,0,0,0,1]),
  ...Array.from({length:20}, () => [0,0,0,0,0,0,1])
]}));
\end{verbatim}
\textbf{Python executed immediately after the action}
\begin{verbatim}
import json
x = json.load(open(
    'benchmark_inputs/current_observation/observation.json'))
print(x['state'])
print(x['proprioception'])
\end{verbatim}
\hrule
\vspace{6pt}
\textbf{Observed values after each lift-and-hold batch}
\vspace{5pt}

\centering
\begin{tabular}{@{}lccc@{}}
\toprule
Probe & Sensor $F_z$ (N) & EEF height (m) & Jaw width (m) \\
\midrule
Left mug & $-8.635$ & 0.207 & 0.0613 \\
Middle mug & $-12.195$ & 0.180 & 0.0622 \\
Right mug & $-6.861$ & 0.204 & 0.0616 \\
\bottomrule
\end{tabular}
\par\vspace{6pt}
\raggedright
\textbf{Final evaluation:} all three mugs sampled; heaviest mug in
the green bin, released and clear of the gripper; 4/4 events completed.
\vspace{6pt}
\hrule
\end{minipage}
\caption{Active sensing in a successful weighing episode. The same
lift-and-hold command pattern is issued for each mug, followed by a
state and proprioception read. Values are rounded from the returned
observations. Mug locations refer to the initial head-camera view;
height is in the robot-base frame and force is in the sensor frame.}
\label{fig:app-weighing-program}
\end{figure}

\clearpage

\section{Per-task Results Across Three Rollouts}
\label{app:task-statistics}

In this section, we record the per task rollout results. Binary entries give full-task successes out of three rollouts.
Long-horizon entries give the sum of three normalized stage-completion
fractions divided by three; $K$ denotes the number of stages.
The supplementary material provides the complete rollout-level evaluation records.

\begin{table}[htbp]
\centering
\caption{Perception: full-task successes out of three rollouts.}
\label{tab:app-task-perception}
\scriptsize
\setlength{\tabcolsep}{1.2pt}
\renewcommand{\arraystretch}{1.25}
\begin{tabular*}{\linewidth}{@{\extracolsep{\fill}}l*{10}{c}@{}}
\toprule
Agent & \shortstack{T01\\Upright\\marker} & \shortstack{T02\\Between\\objects} & \shortstack{T03\\Red\\object} & \shortstack{T04\\Striped\\object} & \shortstack{T05\\Marked\\instance} & \shortstack{T06\\Ball\\shape} & \shortstack{T07\\Cola\\drink} & \shortstack{T08\\Heaviest\\mug} & \shortstack{T09\\Highest\\reflectance} & \shortstack{T10\\Largest\\cylinder} \\
\midrule
GPT-6 Astra & $3/3$ & $3/3$ & $3/3$ & $3/3$ & $3/3$ & $3/3$ & $3/3$ & $3/3$ & $3/3$ & $3/3$ \\
GPT-5.6 Sol & $1/3$ & $0/3$ & $2/3$ & $1/3$ & $2/3$ & $3/3$ & $1/3$ & $1/3$ & $1/3$ & $2/3$ \\
Claude Opus 5 & $3/3$ & $2/3$ & $3/3$ & $3/3$ & $3/3$ & $3/3$ & $3/3$ & $2/3$ & $3/3$ & $3/3$ \\
Claude Fable 5.1 & $2/3$ & $2/3$ & $1/3$ & $3/3$ & $3/3$ & $3/3$ & $3/3$ & $2/3$ & $3/3$ & $3/3$ \\
DeepSeek V4.1 Flash & $1/3$ & $1/3$ & $1/3$ & $1/3$ & $0/3$ & $1/3$ & $0/3$ & $1/3$ & $1/3$ & $2/3$ \\
Kimi K3 & $1/3$ & $0/3$ & $0/3$ & $2/3$ & $0/3$ & $3/3$ & $1/3$ & $1/3$ & $1/3$ & $2/3$ \\
Qwen3.8-Max & $1/3$ & $2/3$ & $2/3$ & $2/3$ & $2/3$ & $3/3$ & $1/3$ & $1/3$ & $3/3$ & $3/3$ \\
\bottomrule
\end{tabular*}
\end{table}

\begin{table}[htbp]
\centering
\caption{Short-horizon manipulation: full-task successes out of three rollouts.}
\label{tab:app-task-manipulation}
\scriptsize
\setlength{\tabcolsep}{1.2pt}
\renewcommand{\arraystretch}{1.25}
\begin{tabular*}{\linewidth}{@{\extracolsep{\fill}}l*{10}{c}@{}}
\toprule
Agent & \shortstack{T11\\Pick\\soup} & \shortstack{T12\\Push\\plate} & \shortstack{T13\\Press\\button} & \shortstack{T14\\Close\\drawer} & \shortstack{T15\\Open\\drawer} & \shortstack{T16\\Open\\microwave} & \shortstack{T17\\Turn on\\stove} & \shortstack{T18\\Insert\\tube} & \shortstack{T19\\Wipe\\spill} & \shortstack{T20\\Pour\\wine} \\
\midrule
GPT-6 Astra & $3/3$ & $3/3$ & $3/3$ & $3/3$ & $3/3$ & $3/3$ & $2/3$ & $0/3$ & $2/3$ & $3/3$ \\
GPT-5.6 Sol & $3/3$ & $0/3$ & $2/3$ & $3/3$ & $0/3$ & $0/3$ & $2/3$ & $0/3$ & $0/3$ & $2/3$ \\
Claude Opus 5 & $3/3$ & $1/3$ & $3/3$ & $2/3$ & $3/3$ & $0/3$ & $1/3$ & $0/3$ & $1/3$ & $2/3$ \\
Claude Fable 5.1 & $3/3$ & $0/3$ & $3/3$ & $2/3$ & $1/3$ & $0/3$ & $1/3$ & $0/3$ & $1/3$ & $2/3$ \\
DeepSeek V4.1 Flash & $1/3$ & $0/3$ & $1/3$ & $0/3$ & $0/3$ & $0/3$ & $1/3$ & $0/3$ & $1/3$ & $1/3$ \\
Kimi K3 & $2/3$ & $0/3$ & $1/3$ & $2/3$ & $1/3$ & $0/3$ & $1/3$ & $0/3$ & $1/3$ & $1/3$ \\
Qwen3.8-Max & $2/3$ & $0/3$ & $3/3$ & $2/3$ & $2/3$ & $0/3$ & $0/3$ & $0/3$ & $1/3$ & $1/3$ \\
\bottomrule
\end{tabular*}
\end{table}

\begin{table}[htbp]
\centering
\caption{Long-horizon manipulation: summed normalized stage-completion scores divided by three.}
\label{tab:app-task-long-horizon}
\scriptsize
\setlength{\tabcolsep}{1.2pt}
\renewcommand{\arraystretch}{1.25}
\begin{tabular*}{\linewidth}{@{\extracolsep{\fill}}l*{10}{c}@{}}
\toprule
Agent & \shortstack{T21\\Cooking\\(text)} & \shortstack{T22\\Cooking\\(image)} & \shortstack{T23\\Drawer\\search} & \shortstack{T24\\Pour\\twice} & \shortstack{T25\\Rotate\\bowls} & \shortstack{T26\\Cabinet\\transfer} & \shortstack{T27\\Drawer\\storage} & \shortstack{T28\\Unlatch\\door} & \shortstack{T29\\Hammer\\nail} & \shortstack{T30\\Use\\seesaw} \\
$K$ & 5 & 5 & 8 & 4 & 4 & 3 & 4 & 2 & 3 & 4 \\
\midrule
GPT-6 Astra & $1.80/3$ & $0.80/3$ & $3.00/3$ & $2.00/3$ & $0.00/3$ & $3.00/3$ & $1.00/3$ & $2.50/3$ & $2.00/3$ & $1.00/3$ \\
GPT-5.6 Sol & $0.80/3$ & $0.20/3$ & $0.25/3$ & $2.00/3$ & $0.75/3$ & $1.67/3$ & $1.00/3$ & $0.00/3$ & $0.67/3$ & $0.00/3$ \\
Claude Opus 5 & $0.00/3$ & $0.40/3$ & $0.00/3$ & $1.50/3$ & $0.00/3$ & $0.33/3$ & $0.00/3$ & $0.50/3$ & $1.00/3$ & $0.00/3$ \\
Claude Fable 5.1 & $0.20/3$ & $0.20/3$ & $0.00/3$ & $1.50/3$ & $0.00/3$ & $0.00/3$ & $0.00/3$ & $1.00/3$ & $2.33/3$ & $0.00/3$ \\
DeepSeek V4.1 Flash & $0.00/3$ & $0.80/3$ & $0.00/3$ & $0.75/3$ & $0.00/3$ & $1.33/3$ & $0.00/3$ & $0.00/3$ & $1.00/3$ & $0.00/3$ \\
Kimi K3 & $0.00/3$ & $0.80/3$ & $1.00/3$ & $0.75/3$ & $0.00/3$ & $0.00/3$ & $0.50/3$ & $0.00/3$ & $0.33/3$ & $0.00/3$ \\
Qwen3.8-Max & $0.80/3$ & $0.40/3$ & $0.62/3$ & $2.00/3$ & $0.00/3$ & $0.67/3$ & $0.00/3$ & $0.00/3$ & $0.00/3$ & $0.00/3$ \\
\bottomrule
\end{tabular*}
\end{table}

\clearpage

\begin{table}[htbp]
\centering
\caption{Observation ablation: task successes out of three rollouts (L1--L3).}
\label{tab:app-task-observation}
\scriptsize
\setlength{\tabcolsep}{1.0pt}
\renewcommand{\arraystretch}{1.25}
\begin{tabular*}{\linewidth}{@{\extracolsep{\fill}}ll*{10}{c}@{}}
\toprule
Agent & Obs. & \shortstack{T11\\Pick\\soup} & \shortstack{T12\\Push\\plate} & \shortstack{T13\\Press\\button} & \shortstack{T14\\Close\\drawer} & \shortstack{T15\\Open\\drawer} & \shortstack{T16\\Open\\microwave} & \shortstack{T17\\Turn on\\stove} & \shortstack{T18\\Insert\\tube} & \shortstack{T19\\Wipe\\spill} & \shortstack{T20\\Pour\\wine} \\
\midrule
Astra & L1 & $2/3$ & $2/3$ & $1/3$ & $2/3$ & $3/3$ & $3/3$ & $2/3$ & $0/3$ & $1/3$ & $3/3$ \\
 & L2 & $3/3$ & $3/3$ & $3/3$ & $2/3$ & $3/3$ & $2/3$ & $1/3$ & $0/3$ & $2/3$ & $3/3$ \\
 & L3 & $3/3$ & $3/3$ & $3/3$ & $3/3$ & $3/3$ & $3/3$ & $2/3$ & $0/3$ & $2/3$ & $3/3$ \\
\addlinespace[3pt]
Opus & L1 & $1/3$ & $0/3$ & $3/3$ & $2/3$ & $0/3$ & $0/3$ & $0/3$ & $0/3$ & $0/3$ & $0/3$ \\
 & L2 & $0/3$ & $1/3$ & $3/3$ & $0/3$ & $0/3$ & $0/3$ & $0/3$ & $0/3$ & $0/3$ & $0/3$ \\
 & L3 & $3/3$ & $1/3$ & $3/3$ & $2/3$ & $3/3$ & $0/3$ & $1/3$ & $0/3$ & $1/3$ & $2/3$ \\
\addlinespace[3pt]
Fable & L1 & $1/3$ & $0/3$ & $1/3$ & $0/3$ & $0/3$ & $0/3$ & $0/3$ & $0/3$ & $0/3$ & $0/3$ \\
 & L2 & $1/3$ & $0/3$ & $2/3$ & $1/3$ & $0/3$ & $0/3$ & $1/3$ & $0/3$ & $0/3$ & $0/3$ \\
 & L3 & $3/3$ & $0/3$ & $3/3$ & $2/3$ & $1/3$ & $0/3$ & $1/3$ & $0/3$ & $1/3$ & $2/3$ \\
\bottomrule
\end{tabular*}
\end{table}

\begin{table}[htbp]
\centering
\caption{Context ablation: summed normalized stage-completion scores divided by three.}
\label{tab:app-task-context}
\scriptsize
\setlength{\tabcolsep}{1.0pt}
\renewcommand{\arraystretch}{1.25}
\begin{tabular*}{\linewidth}{@{\extracolsep{\fill}}ll*{10}{c}@{}}
\toprule
Agent & Context & \shortstack{T21\\Cooking\\(text)} & \shortstack{T22\\Cooking\\(image)} & \shortstack{T23\\Drawer\\search} & \shortstack{T24\\Pour\\twice} & \shortstack{T25\\Rotate\\bowls} & \shortstack{T26\\Cabinet\\transfer} & \shortstack{T27\\Drawer\\storage} & \shortstack{T28\\Unlatch\\door} & \shortstack{T29\\Hammer\\nail} & \shortstack{T30\\Use\\seesaw} \\
\midrule
Astra & None & $1.80/3$ & $0.80/3$ & $3.00/3$ & $2.00/3$ & $0.00/3$ & $3.00/3$ & $1.00/3$ & $2.50/3$ & $2.00/3$ & $1.00/3$ \\
 & Video & $1.60/3$ & $1.60/3$ & $3.00/3$ & $3.00/3$ & $3.00/3$ & $2.67/3$ & $3.00/3$ & $2.50/3$ & $3.00/3$ & $2.25/3$ \\
 & Video + Traj. & $2.40/3$ & $1.40/3$ & $3.00/3$ & $2.50/3$ & $2.00/3$ & $3.00/3$ & $3.00/3$ & $1.50/3$ & $3.00/3$ & $2.00/3$ \\
\addlinespace[3pt]
Opus & None & $0.00/3$ & $0.40/3$ & $0.00/3$ & $1.50/3$ & $0.00/3$ & $0.33/3$ & $0.00/3$ & $0.50/3$ & $1.00/3$ & $0.00/3$ \\
 & Video & $0.40/3$ & $0.00/3$ & $0.12/3$ & $1.50/3$ & $0.75/3$ & $1.33/3$ & $0.50/3$ & $0.00/3$ & $2.00/3$ & $0.00/3$ \\
 & Video + Traj. & $0.20/3$ & $0.20/3$ & $1.00/3$ & $1.50/3$ & $0.00/3$ & $1.00/3$ & $0.75/3$ & $0.00/3$ & $2.67/3$ & $0.00/3$ \\
\addlinespace[3pt]
Fable & None & $0.20/3$ & $0.20/3$ & $0.00/3$ & $1.50/3$ & $0.00/3$ & $0.00/3$ & $0.00/3$ & $1.00/3$ & $2.33/3$ & $0.00/3$ \\
 & Video & $0.20/3$ & $0.00/3$ & $1.38/3$ & $1.50/3$ & $1.00/3$ & $0.33/3$ & $0.50/3$ & $0.00/3$ & $3.00/3$ & $0.25/3$ \\
 & Video + Traj. & $0.00/3$ & $0.00/3$ & $1.25/3$ & $2.00/3$ & $1.25/3$ & $0.67/3$ & $0.75/3$ & $0.00/3$ & $2.33/3$ & $0.75/3$ \\
\bottomrule
\end{tabular*}
\end{table}

\clearpage

\section{Qualitative Rollout Visualizations}
\label{app:rollout-visualizations}

We visualize successful and unsuccessful executions from the perception,
short-horizon manipulation, and long-horizon suites.
Each example shows synchronized head-camera and wrist-camera views at selected
control steps. 
The perception examples share the same task, while the manipulation and
long-horizon examples illustrate different tasks within their respective suites.

\subsection{Perception}
\label{app:rollout-perception}

\begin{figure}[htbp]
\centering
\includegraphics[width=\linewidth]{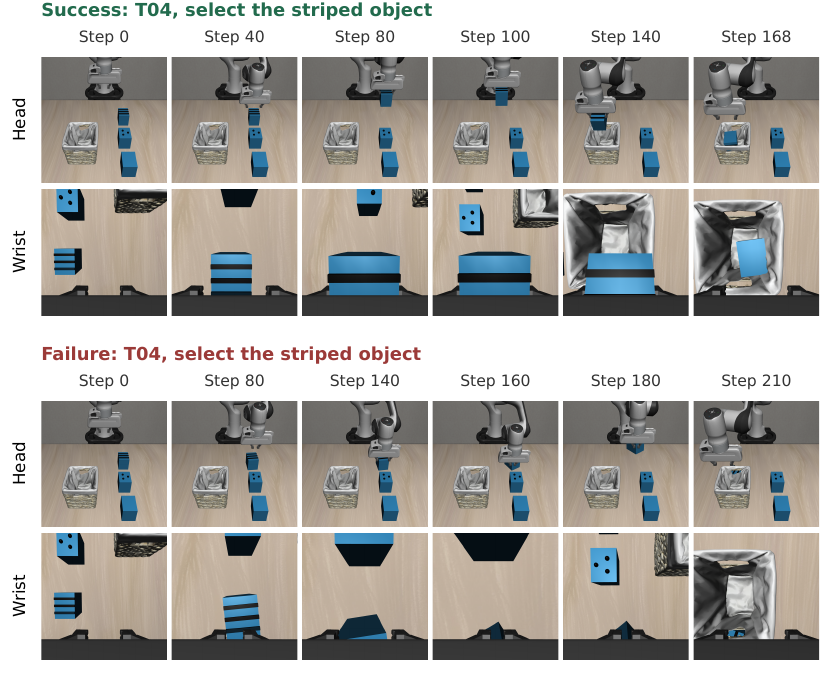}
\caption{\textbf{Perception rollouts on T04.}
The instruction is to place the striped object in the collection bin.
The successful execution transfers the striped object to the bin.
In the unsuccessful execution, the robot also transports the striped object
toward the bin, but its final placement does not satisfy the task's success check.
The wrist view exposes the texture cues and the objects near the gripper,
while the head view shows their positions relative to the bin.}
\label{fig:app-rollout-perception}
\end{figure}

\clearpage
\subsection{Short-horizon manipulation}
\label{app:rollout-manipulation}

\begin{figure}[htbp]
\centering
\includegraphics[width=\linewidth]{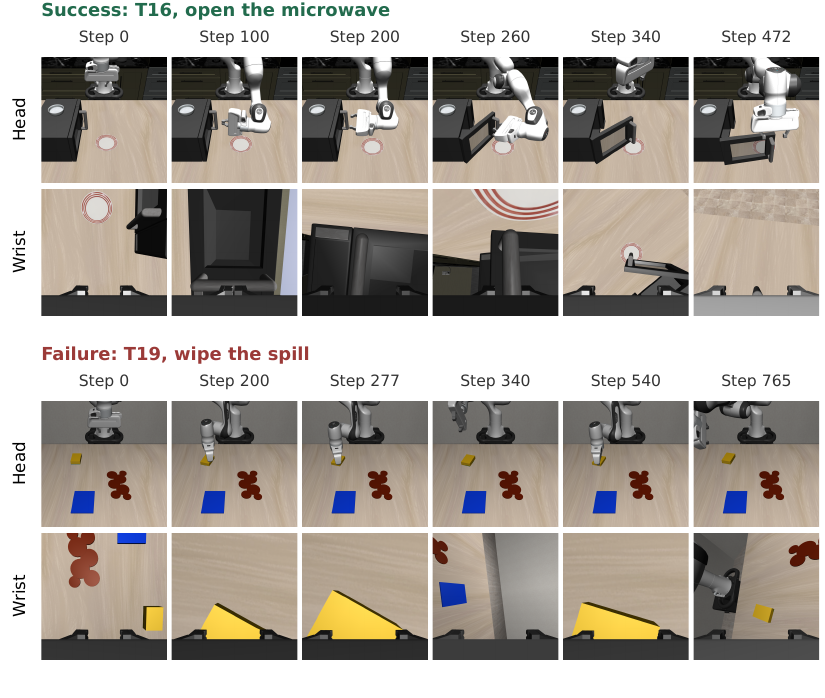}
\caption{\textbf{Short-horizon manipulation rollouts.}
\emph{Top:} a successful T16 execution opens the microwave door.
The selected frames show approach, gripper reorientation, and door opening.
\emph{Bottom:} an unsuccessful T19 execution attempts to grasp the sponge
but leaves the spill unwiped when the time budget expires.
The brown spill remains visible throughout the selected frames.
The paired views show the gripper configuration and its relation to the
handle or sponge.}
\label{fig:app-rollout-manipulation}
\end{figure}

\clearpage
\subsection{Long-horizon execution}
\label{app:rollout-long-horizon}

\begin{figure}[htbp]
\centering
\includegraphics[width=\linewidth]{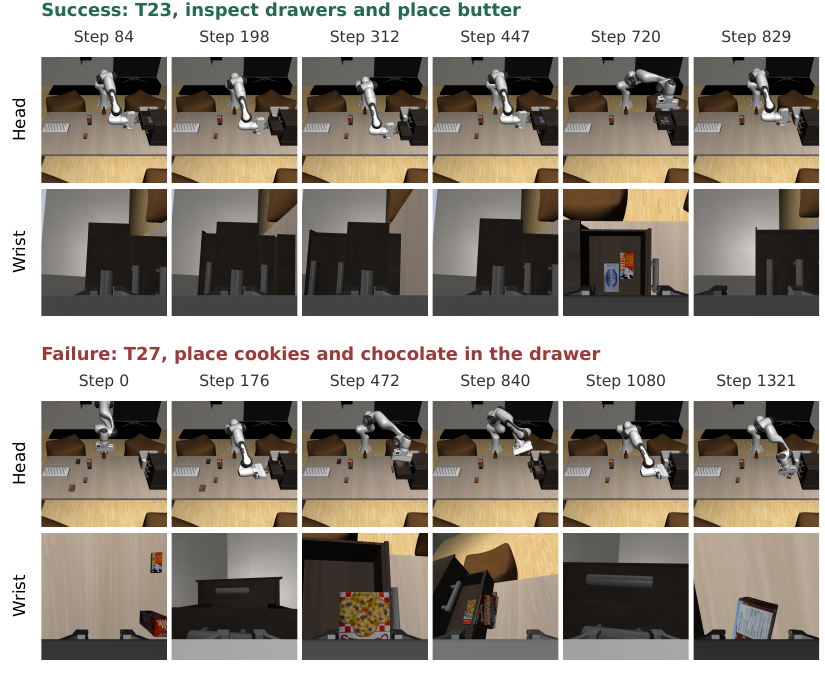}
\caption{\textbf{Long-horizon rollouts.}
\emph{Top:} a successful T23 execution opens and closes the drawers to inspect
their contents, reopens the occupied drawer, and places butter inside.
The selected frames show the top, middle, and bottom drawer openings,
the occupied drawer reopened, butter placement, and the final configuration.
All eight required stages are completed.
\emph{Bottom:} an unsuccessful T27 execution opens the middle drawer and
places cookies inside, but does not complete the subsequent chocolate
placement before the time budget expires. It completes two of four required
stages; the final frame shows the chocolate outside the drawer.}
\label{fig:app-rollout-long-horizon}
\end{figure}

\end{document}